\documentclass[letterpaper, 10 pt, conference]{ieeeconf}  % Comment this line out if you need a4paper

\IEEEoverridecommandlockouts                              % This command is only needed if 
\usepackage{graphicx}
\graphicspath{{./figs/}}
\usepackage{amsmath}

\usepackage{etex} %%% pour allouer + de place pour les variables -
\usepackage[utf8]{inputenc}
\usepackage[T1]{fontenc}
\usepackage{fixltx2e}
\usepackage{longtable}
\usepackage{float}
\usepackage{wrapfig}
\usepackage{soul}
\usepackage{textcomp}
\usepackage{marvosym}
\usepackage{wasysym}
\usepackage{latexsym}
\usepackage{hyperref}
\usepackage[table]{xcolor}
\usepackage{dcolumn}
\usepackage{booktabs}
\usepackage{subcaption} %%% for using subtable environment
\usepackage[labelformat=simple]{subcaption}

\usepackage{enumerate} %%

\usepackage{graphicx}
\usepackage{tikz}
\usetikzlibrary{decorations.fractals}

\usepackage{empheq}
\usepackage{bm} %% for bold math fonts
\usepackage{movie15}

\usepackage{amsfonts}
\usepackage{amscd}
\usepackage{amssymb}
\usepackage[export]{adjustbox}%% adjust the alignement of the figures

\usepackage[d]{esvect}
\usepackage{manfnt} %% pour fontes $\textdbend$
\usepackage{xcolor}
\newcommand{\Vect}[1]{%
  \mathbf{#1}
}

\usepackage{amsmath,mleftright}
\usepackage{xparse}
\NewDocumentCommand{\evalat}{sO{\big}mm}{%
  \IfBooleanTF{#1}
   {\mleft. #3 \mright|_{#4}}
   {#3#2|_{#4}}%
}

\title{\LARGE \bf
  An optimization framework for simulation and kinematic control of Constrained Collaborative Mobile Agents (CCMA) system
}

\author{Nitish Kumar$^{*}$, Stelian Coros
\thanks{Nitish Kumar and Stelian Coros are with the Computational Robotics Lab at the Institute for Pervasive Computing, ETH Zurich Switzerland. $\{$nitish.kumar@inf.ethz.ch, stelian.coros@inf.ethz.ch$\}$}% <-this % stops a space
\thanks{* corresponding author}
}

\begin{document}

\maketitle
\thispagestyle{empty}
\pagestyle{empty}

%%%%%%%%%%%%%%%%%%%%%%%%%%%%%%%%%%%%%%%%%%%%%%%%%%%%%%%%%%%%%%%%%%%%%%%%%%%%%%%%
\begin{abstract}
%Abstract limit of 200 words
We present a concept of constrained collaborative mobile agents (CCMA) system, which consists of multiple wheeled mobile agents constrained by a passive kinematic chain. This mobile robotic system
is modular in nature, the passive kinematic chain can be easily
replaced with different designs and morphologies for different
functions and task adaptability. Depending solely on the actuation
of the mobile agents, this mobile robotic system can manipulate or
position an end-effector. However, the complexity of the system
due to presence of several mobile agents, passivity of the kinematic chain and the nature of the constrained collaborative manipulation requires development of an optimization framework. 
%for the mobile robotic system to execute its tasks. 
We therefore present an optimization framework
for forward simulation and kinematic control
of this system. With this optimization framework, the number of deployed mobile agents, actuation schemes, the design and morphology of the passive kinematic chain can be easily changed, which reinforces the modularity and collaborative aspects of the mobile robotic system. We present results, in simulation, for spatial $4$-DOF to $6$-DOF CCMA system examples. Finally, we present experimental quantitative results for two different fabricated $4$-DOF prototypes, which demonstrate different actuation schemes, control and collaborative manipulation of an end-effector.
\end{abstract}

\begin{keywords}
  Collaborative robots, Multi-robot systems, Mobile manipulation, Simulation, modeling and control
\end{keywords}

%%%%%%%%%%%%%%%%%%%%%%%%%%%%%%%%%%%%%%%%%%%%%%%%%%%%%%%%%%%%%%%%%%%%%%%%%%%%%%%% 
\section{Introduction}
From structured warehouses to unstructured environments such as construction sites, agricultural fields, research into large scale mobile robotics systems is fuelled by demands for increased mobility, task adaptability while able to perform various functions.  {Most of the mobile robotic systems have a  monolithic hardware architecture consisting of  fully actuated serial manipulator/s mounted on a single mobile base}. 
%Such systems are highly overdesigned, for e.g. an extremely heavy mobile base supporting the weight of a heavy industrial manipulator. 
Due to the monolithic and inflexible hardware architecture, current systems miss the potential advantages of task-adaptability, shared collaborative manipulation brought upon by a more modular, lightweight architecture. 
%Moreover, when certain tasks require higher mobility, such systems might become infeasible because there might not be existing industrial solutions and a system meeting such requirements could be practically infeasible.

\begin{figure}[!t] 
\vspace{7pt}
  \centering
    \includegraphics[width=0.95\columnwidth]{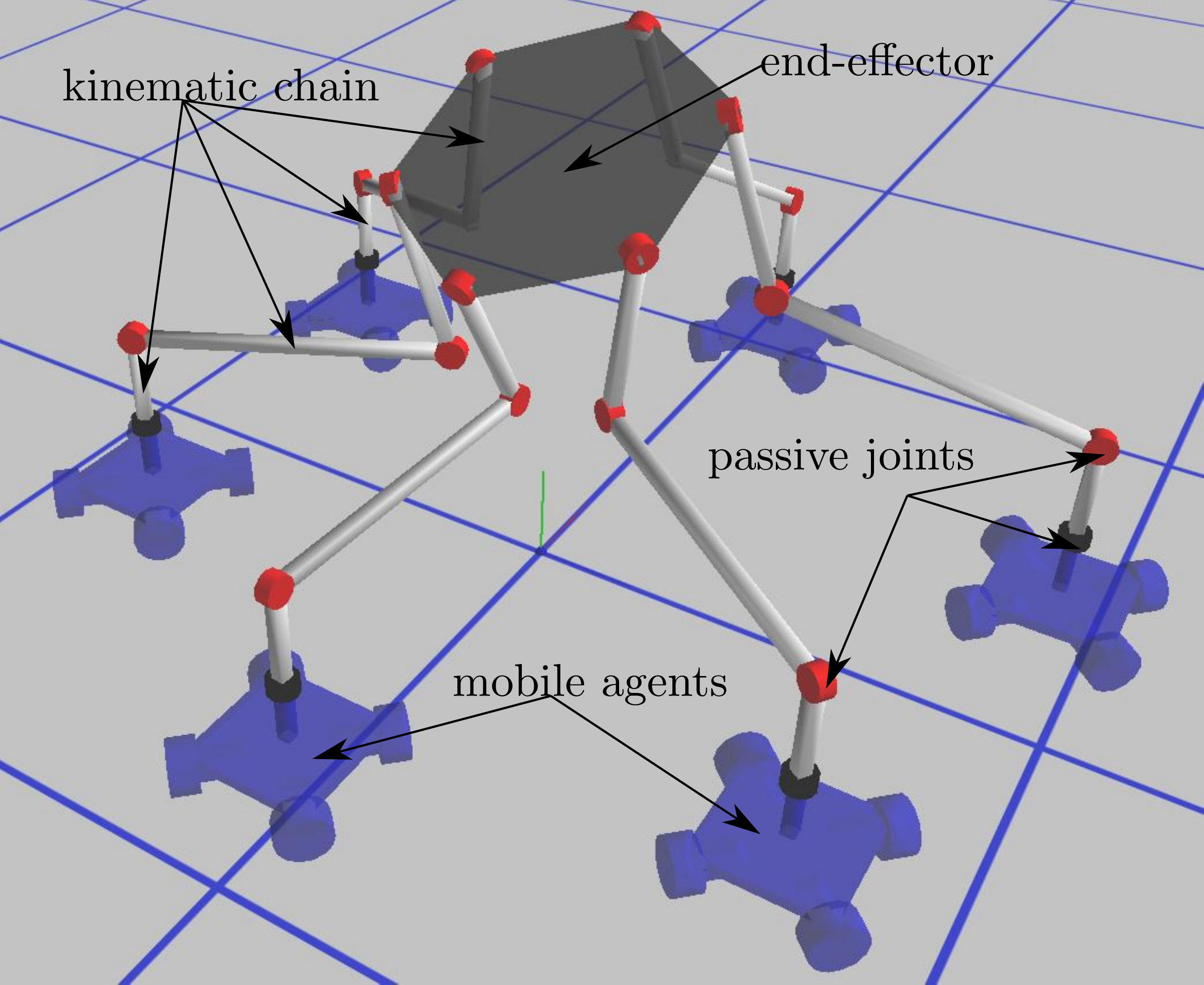}
    \caption{Constrained collaborative mobile agents concept: a number of mobile bases $0,1\cdots n_m-1$ constrained with a passive kinematic chain (gray bars connected with red passive joints) manipulating an end-effector (black polygon); with an option to add an extra joint (first joint in black color) at the base of the mobile robot.}
    \label{fig:concepCCMA}
\end{figure}
%\begin{figure}[!t] 
%  \centering
%  \vspace{7pt}
%    \includegraphics[width=1.0\columnwidth]{concepCCMA}
%    \caption{Examples of a spatial CCMA systems with different number of active mobile bases, different designs and morphologies of the passive kinematic chain.}
%    \label{fig:concepCCMA}
%\end{figure}

Therefore, we present an alternate concept for a modular, collaborative and task-adaptable mobile robotic system consisting of multiple mobile bases (Fig.~\ref{fig:concepCCMA}). Our goal is to develop a unified system that allows us to explore 1) collaborative manipulation using teams of mobile robots, and 2) reconfigurable manipulators that use simple robotic systems as building blocks. 
%The CCMA system presented in the current work, consists of multiple mobile bases constrained by a passive kinematic chain, thus forming a closed loop mechanism. 
CCMA system is modular in nature such that the number of mobile bases, actuation schemes and the design, morphology of the passive kinematic chain can be easily changed in hardware. The task of a single heavy mobile base is substituted by collaboration between and shared manipulation by several light weight mobile bases. Fully actuated serial manipulator/s are substituted by a fully constrained passive closed-loop kinematic mechanism. 
%This hardware architecture imposes several kinematic constraints on the individual motion of the mobile agents. 
%whose design and morphology can be easily changed depending on the task requirements. 

The above mentioned flexibilities in the hardware architecture can only be realized in practice, if a corresponding flexible software framework is developed for easier and faster evaluation of vast design and actuation space that CCMA system presents. Therefore, we develop and present an optimization framework for simulation and control of CCMA system, which allows for easy and fast change in actuation schemes, number of mobile agents, design and morphology of the passive kinematic chain. We evaluate this optimization framework on different prototypes of the CCMA system, in simulation and experiments, for end-effector manipulation and positioning tasks.

\subsection{Related Work on System Architecture}

In the field of construction robotics or robotics in architecture, process specific digital fabrication techniques such as robotic brickwork~\cite{dorfler_mobile_2016}, robotic formwork~\cite{kumar_design_2017} increasingly  make use of more generic mobile robotic systems~\cite{loveridge_robots_2017}. These mobile robotic systems typically include a fully actuated industrial manipulator mounted on a single mobile base which can be wheeled or tracked~\cite{giftthaler_mobile_2017,keating_toward_2017}. 
%The current mobile robotic systems are monolithic in hardware architecture 
%As such mobile robotic systems comprising of fully actuated industrial manipulators on a single mobile agent are not scalable and adaptable in hardware to changing task and corresponding mobility requirements. 

{There are several reports of different modular hardware architectures which use multiple mobile agents to collaborative perform a task}. For e.g., multiple flying agents were used for building an architecture scale installation in this work~\cite{augugliaro_flight_2014}. However, flight based agents lack the rigidity and stiffness required for more demanding manipulation tasks.
Another approach consists in using wheeled mobile agents to collaboratively manipulate an object which is suspended through several cables~\cite{rasheed_kinematic_2018,zi_localization_2015}.  
Cable driven object manipulation has the capability to provide the rigidity, stiffness and higher payloads. {However to fulfil these requirements, the topology of the cables encircles the object being manipulated, thereby reducing the available workspace around the end-effector.}
%So though the cable driven systems can be useful for logistics and material handling, they also can not replace conventional mobile robotic systems for variety of in situ digital fabrication tasks, where unencumbered workspace around the end-effector is desired.
 
%The commonality in this hardware system architecture is the use of multiple mobile agents as the only actuators in the system. 
%Flight based mobile agents are suitable for discrete material handling, logistics but they can not replace industrial robots, where end-effector manipulation requires rigidity, stiffness and capability of much higher payloads. Cable driven object manipulation, through actuated mobile agents, has the capability to provide the rigidity, stiffness and higher payloads. However to fulfil these requirements, the topology of the cables encircles the object being manipulated. So though the cable driven systems can be useful for logistics and material handling, they also can not replace the industrial manipulators for variety of in situ digital fabrication tasks such as drilling, robotic brickwork~\cite{dorfler_mobile_2016}, robotic formwork~\cite{kumar_design_2017} where unencumbered workspace around the end-effector is desired. 

{Hardware system architecture, where a passive kinematic chain with rigid links is connected to multiple mobile agents (CCMA), has the potential to replace conventional large scale mobile robotic systems for a variety of tasks. Owing to higher rigidity of the links and flexible arrangement of the kinematic chain around the end-effector, such systems would allow for open workspace around the end-effector to carry out process-specific tasks}. Few prototypes with this system architecture have been previously reported in the work~\cite{wan_design_2010, hu_singularity_2009, hu_type_2009}. However, morphology of the passive kinematic chain was limited to $6$-DOF, with a single link with two passive joints connecting the end-effector to each mobile base. Choice of the actuation scheme was conservative and a fixed number of mobile agents were used. 
This misses a big potential advantage, where over-actuation along with use of multiple mobile agents and different actuation schemes can be exploited to enhance the system performance.

{In the current work, we present a generalized concept for this hardware system architecture}. Our work accommodates a far richer design space for the passive kinematic chains and diverse actuation schemes for multiple mobile agents.

\subsection{Related Work on Simulation, Modeling and Kinematic Control Techniques}

An overview of kinematic modeling techniques for robotic systems with closed loop kinematic chains can be found in~\cite{murray_mathematical_1994, featherstone_rigid_2008, sciavicco_modelling_2000, merlet_parallel_2006}. These techniques result in models in the form of $\Vect{J_x}\cdot \dot{\Vect{X}} = \Vect{J_q}\cdot\dot{\Vect{q}}$, either obtained through differentiating forward position models of the form $\Vect{X}= f(\Vect{q})$ or directly through screw theory~\cite{davidson_robots_2004} based generation of twist and wrench systems~\cite{murray_mathematical_1994, merlet_parallel_2006}. $\Vect{X}$ represents the end-effector variables and $\Vect{q}$ contains the independent active control variables, such as motor angles. These two $\Vect{J_x}$ and $\Vect{J_q}$ matrices form the essential components for numerical techniques either for solving forward kinematics problems (simulation) or inverse kinematics problems~\cite{whitney_resolved_1969} (control) through the relation $\Vect{J_x}\cdot \delta \Vect{X} = \Vect{J_q} \cdot \delta \Vect{q}$.
%\begin{equation}
%\Vect{J_x}\cdot \delta \Vect{X} = \Vect{J_q} \cdot \delta \Vect{q}
%\end{equation}

%\textcolor{red}{Even though the above numerical methods and modeling techniques are general and lead to fast computation of forward and inverse kinematic solutions. The implementation of the Jacobian matrices $\Vect{J_x}$ and $\Vect{J_q}$ is still robot specific with manual error prone time consuming steps, like differentiation of robotic specific forward position models, elimination steps for closed loop kinematic chains. Due to these reasons, Jacobian based methods are practically infeasible when tasked with evaluating a class of robots, where the robot space is vast, as is the case for the CCMA system. However, optimization-based control scheme proposed in the current work is very generic, and it is readily applicable to any instance of a CCMA system without requiring any new code to be written, or any new Jacobian terms to be derived. In other words, we have developed a modular control scheme to accompany the novel concept of modular CCMA robotic systems.}
%The jacobian matrices $\Vect{J_x}$ and $\Vect{J_q}$ still have to be generated manually for each robot, which is time consuming especially for robotic systems with closed loop kinematic chains. Moreover, it involves manual error prone steps like differentiation of forward position models, elimination steps in case of systems with closed loop kinematic chains to remove passive joints from $\Vect{q}$. 
{Even though the above numerical methods and modeling techniques lead to fast computation of forward and inverse kinematic solutions, the implementation of the Jacobian matrices $\Vect{J_x}$ and $\Vect{J_q}$ is still tailored to individual robots with manual error prone time consuming steps, like differentiation of robot specific forward position models, elimination steps for closed loop kinematic chains. 
%Due to these reasons, Jacobian based methods are practically infeasible when tasked with evaluating a class of robots, where the robot space is vast, as is the case for the CCMA system. 
However, optimization-based simulation and control scheme proposed in the current work is very generic, and it is readily applicable to any instance of a CCMA system without requiring any new code to be written, or any new Jacobian terms to be derived. We have developed a modular simulation and control scheme to accompany the novel concept of modular CCMA systems.}

In the current work, rigid body kinematics is modeled on a constraint based formulation presented in the paper~\cite{coros_computational_2013,thomaszewski_computational_2014}, which abstracts a rigid body robotic system to a collection of rigid bodies connected with kinematic constraints imposed by the joints and actuators. 
%This modeling approach does not involve computation of either $\Vect{J_x}$ or $\Vect{J_q}$ described above. 
We extend this modeling framework by including additional constraints for the mobile agents acting as actuators.
%on top of the constraints imposed by passive joint connections in the closed loop kinematic chain. 
Due to this abstraction, no separate specific implementation is needed for different actuation schemes, number of mobile agents and different designs, morphologies of the passive kinematic chain. 

For the simulation (forward kinematics) of the CCMA system, we calculate the derivatives of the kinematic constraints including those imposed by the actuation of mobile agents analytically. Moreover, we calculate the derivatives of the tasks, formulated as objective functions, with respect to control parameters of the CCMA system for solving the kinematic control problem (inverse kinematics). These derivatives are required for the gradient-based methods (e.g. L-BFGS, Gauss-Newton). For analytical formulation of the derivatives, we utilize the first order sensitivity analysis techniques~\cite{mcnamara_fluid_2004, auzinger_computational_2018,cao_adjoint_2002}. 
%Sensitivity analysis techniques are utilized for optimization problems where closed form solutions of the derivatives does not exist, in general. 
The analytical derivation of the derivatives, as compared to using finite differences, allows for real time computation of both forward kinematics (simulation) and inverse kinematics (kinematic control) of the CCMA system, despite being system independent.

\subsection{Contributions}
Our long term goal is to leverage the advantages brought by combining robotic mobility and manipulation capabilities. To this end, in this paper we show that driven by appropriate control systems, very simple robots equipped with very simple manipulators can be quite dexterous. The mobile manipulation systems, we study in this work, have much greater workspace and reach than stationary robots. They are very lightweight, and they can easily be reconfigured to fit the needs of different tasks.
In this paper, we make following contributions: 

\begin{itemize}
\item A general concept for constrained collaborative mobile agents (CCMA), in which an end-effector is connected via a passive kinematic chain to multiple mobile agents.
\item Varying topology of the kinematic chain and actuation schemes of the mobile agents for task adaptability.
\item A unified optimization framework for simulation and kinematic control of CCMA systems, independent of design, morphology of the passive kinematic chain and actuation schemes, number of mobile agents.
\item Two prototypes demonstrating the optimization driven kinematic control of the CCMA system in experiments.
\end{itemize}

%\subsection{Outline}
%
%In section~\ref{sec:ccmaSystem}, we discuss the CCMA concept and the system modeling in detail. The optimization framework for simulation and kinematic control of CCMA is described in section~\ref{sec:model_opti}. Fabricated prototypes and experimental results are presented in section~\ref{sec:results} and their kinematic control utilizing the optimization framework is demonstrated. Finally, we discuss the challenges faced in the current work and the directions for the future work in section~\ref{sec:Conclusion}.

%large scale robotic systems  industrial manipulators Construction robotics
%low payload to weight ratio
%Industrial robots mounted on heavy mobile bases
%High mobility, reachability while keeping systems in check
%Modularity, easily reconfigurable and adaptable to individual tasks
%An optimization framework which allows for modeling  and kinematic control of such systems 
%irrespective of topology and choice of actuation scheme. 

\section{Constrained collaborative mobile agents system description and modeling}
\label{sec:ccmaSystem}
\begin{figure}[!b] 
%\vspace{7pt}
  \centering
    \includegraphics[width=1.0\columnwidth]{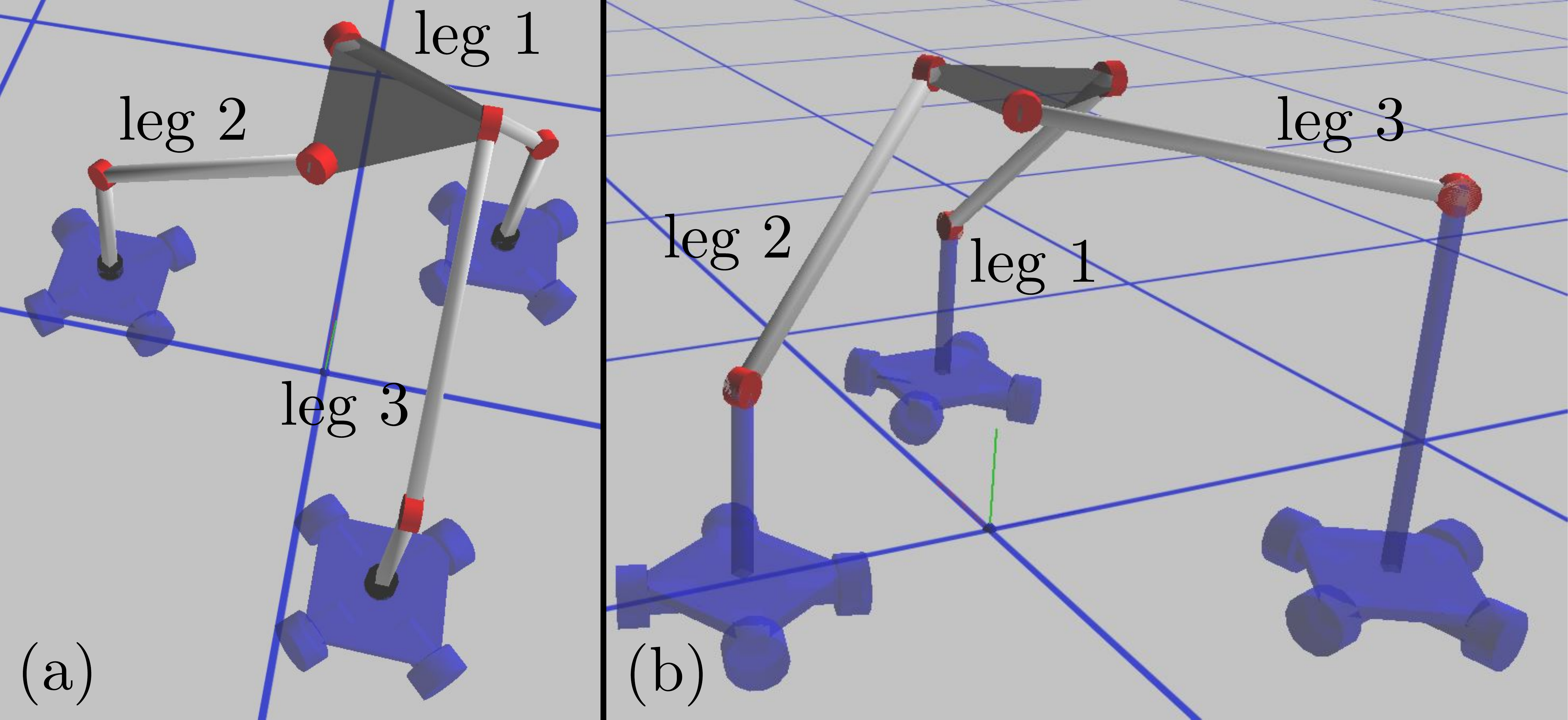}
    \caption{CCMA $4$-DOF examples (a) each leg of the passive kinematic chain has same design (kinematic parameters like length, offset, angles) and morphology); has extra black revolute joints and utilizes the \textit{reduced actuation scheme} (b) with same morphology but different design parameters for leg $3$;  does not have extra black revolute joints and utilizes complete actuation scheme.}
    \label{fig:4DOFExamples}
\end{figure}

\begin{figure}[!t] 
\vspace{7pt}
  \centering
    \includegraphics[width=1.0\columnwidth]{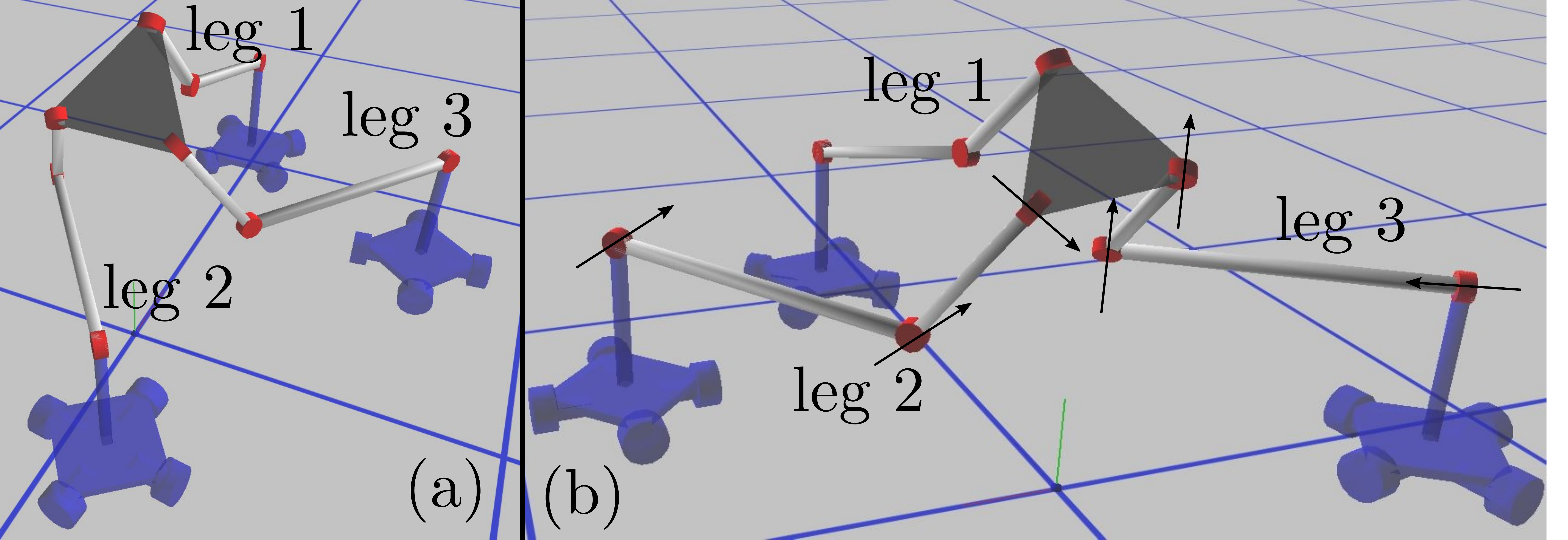}
    \caption{CCMA $6$-DOF (a) symmetric example; each leg of the passive kinematic chain has same design (kinematic parameters like length, offset, angles) and morphology (b) asymmetric example; arrows represent the direction of the revolute joints axes; leg $1$ and leg $2$ have same morphology but different link lengths; leg $3$ has different morphology than the leg $1$ and leg $2$.}
    \label{fig:6DOFExamples}
\end{figure}

{Fig.~\ref{fig:concepCCMA} illustrates the concept of CCMA system. It consists of a passive kinematic chain connecting the end-effector (black polygon)  to a number of omni-directional mobile bases either through a fixed connection or an extra revolute joint (in black color) at the mobile base. The mobile agents form the only actuators in the system to control the end-effector. %The mobile agents can have DOF ranging from $2$ (tracked or wheeled mobile robots with non-holonomic constraints) to $6$ (quadruped robots).  
%In the current paper, we will focus on an instantiation of this concept, where mobile agents are omni-directional  each having $3$-DOF. 
Omni-directional mobile base has $3$-DOF, each of which is used as an actuator}. 
%For e.g. in Fig.~\ref{fig:concepCCMA}, there are a total of $18$ actuators to control the $6$~DOF of the end-effector. 
The extra passive black revolute joint is along the rotation axis of the omni-directional robot. Because of this revolute joint, actuation due to rotation of the omni-directional base has no effect on the end-effector motion. {This reduces the effective control variables in each omni-directional robot to $2$ (\textit{reduced actuation scheme}) from $3$ (\textit{complete actuation scheme}).} 
%We will call it a \textit{reduced actuation scheme} for the CCMA system. 
%When complete actuation of the omni-directional mobile bases are considered, the extra black revolute joints are not present in the CCMA system.
The CCMA example in Fig.~\ref{fig:4DOFExamples}(a) has black revolute joints, therefore it utilizes effective $6$ control variables to manipulate the $4$-DOF end-effector. {Whereas  the CCMA example in Fig.~\ref{fig:4DOFExamples}(b) has a fixed connection (no black revolute joint), therefore it utilizes effective $9$ control variables.} %to manipulate the $4$-DOF end-effector.
%\begin{figure}[!b] 
%%\vspace{7pt}
%  \centering
%    \includegraphics[width=0.95\columnwidth]{concepCCMA}
%    \caption{Constrained collaborative mobile agents concept: a number of mobile bases $0,1\cdots n_m-1$ constrained with a passive kinematic chain (gray bars connected with red passive joints) manipulating an end-effector (black polygon); with an option to add an extra joint (first joint in black color) at the base of the mobile robot.}
%    \label{fig:concepCCMA}
%\end{figure}
{The number of mobile bases ($6$ in Fig.~\ref{fig:concepCCMA}, $3$ in Fig.~\ref{fig:4DOFExamples} and Fig~\ref{fig:6DOFExamples}) can be varied in the CCMA system.} Moreover, design and morphology of the passive kinematic chain can be varied in the CCMA system, as shown with different design and morphologies in Figs.~\ref{fig:concepCCMA},~\ref{fig:4DOFExamples} and~\ref{fig:6DOFExamples} for different CCMA system examples.
%By adding more mobile bases, number of actuators in the CCMA system can be changed. Also topologies presented in these figures represent small subset of morphologies possible. 
%Mobile bases as actuators mean that the resulting CCMA systems can have very large translational workspaces especially along the ground plane ($\mathbf{x_g}$-$\mathbf{y_g}$ plane) and large orientation workspace along the vertical axis $\mathbf{z_g}$.

%Thus the CCMA system has the potential to be scalable to large scale systems with high mobility. This is in contrast to the parallel robots~\cite{merlet_parallel_2006}, where a similar passive kinematic chain with an end-effector is actuated with motors at the fixed base. For parallel robots, usually the number of actuators matches exactly the number of DOF of the end-effector and they have very limited workspace both in translation and orientation. 
 
%Two examples of this hardware architecture are shown in Fig.~\ref{fig:sim4DOFCCMA} and  Fig.~\ref{fig:sim3DOFCCMA}.
\label{sec:CCMASystemModeling}

\subsection{{Notation and preliminaries}}
%Let us assume that the passive kinematic chain along with the extra actuated joint is either a revolute or prismatic joint.
\begin{description}
\item[$n_b$] the total number of the rigid bodies in the CCMA system inclusive of passive kinematic chain and all the mobile bases.
\item[$n_m$] the number of planar mobile robots.
\item[$\Vect{s}$] state vector of the system, which has a size of $6 \cdot n_b$
\item[$\Vect{u}$] 
%the size of the control vector $\Vect{u}$ depends on the number of control variables in the system. 
{the size of control vector $\Vect{u}$ is $3 \cdot n_m$. For \textit{reduced actuation scheme}, the control variable corresponding to orientation of the mobile base can be arbitrarily fixed leaving effective $2$ control parameters for a single mobile base.} 
%and $2 \cdot n_m$ for \textit{complete actuation scheme} and \textit{reduced actuation scheme}, respectively
%.}
%\item[$\Vect{C}$] a vector of constraints which includes the constraints output by each passive joint (revolute, prismatic, spherical, universal) between the two rigid bodies, motor constraints obtained by fixing the value of actuator in the motorized joint between the two rigid bodies. For more details, please refer to the work in~\cite{coros_computational_2013, thomaszewski_computational_2014}. Additionally in this work, for each planar mobile robot in world reference frame
%($O_g$,$\mathbf{x_g}$,$\mathbf{y_g}$,$\mathbf{z_g}$), we add additional kinematic constraints which correspond to the planar constraint on the mobile robot, allowing only translation along $\mathbf{x_g}$,$\mathbf{y_g}$ and rotation about $\mathbf{z_g}$ and motor constraints assuming two motorized prismatic actuators along $\mathbf{x_g}$,$\mathbf{y_g}$ and a rotary actuator along the free axis $\mathbf{z_g}$. For the \textit{reduced actuation scheme}, rotary actuator along the free axis $\mathbf{z_g}$ is not added. 
\end{description}

{The CCMA system consists of rigid bodies connected with kinematic constraints. 
%Each mobile base is also counted as a rigid body having $6$-DOF with a planar constraint on its motion reducing its DOF to $3$. 
Without any kinematic constraints, each rigid body $i$ has $6$-DOF which is described by its state $\Vect{s_i} = [\gamma_i \quad \beta_i \quad \alpha_i \quad \Vect{T_i}^T]^T$ consisting of three Euler angles and translation vector $\Vect{T_i} = [x_i \quad y_i \quad z_i]^T$ from global reference frame to rigid body local co-ordinate system. Thus any point $\Vect{\bar{p}_i}$ or free vector $\Vect{\bar{v}_i}$ expressed in local co-ordinate system of a rigid body $i$ can be converted to global world co-ordinates as $\Vect{p_i} = \Vect{R_{\gamma_i}} \cdot \Vect{R_{\beta_i}} \cdot \Vect{R_{\alpha_i}} \cdot \Vect{\bar{p}_i} +  \Vect{T_i}$ or $\Vect{v_i} = \Vect{R_{\gamma_i}} \cdot \Vect{R_{\beta_i}} \cdot \Vect{R_{\alpha_i}} \cdot \Vect{\bar{v}_i} $. $\Vect{R}$ is an elementary $3 \times 3$ rotation matrix.}

{The kinematic constraints are imposed between rigid bodies $i$ and $j$, which are collected in a vector of constraints $\Vect{C}$. These kinematic constraints are imposed by the passive kinematic connections in form of joints in the passive kinematic chain and due to actuation of the mobile agents. For example, we illustrate how to formulate the kinematic constraints due to a revolute and a spherical kinematic connection. Furthermore, we also describe how to formulate the kinematic constraints resulting from actuation of the omni-direction mobile bases.}

\subsubsection{Revolute/Spherical joint}
{Let $\Vect{v_i(\Vect{\bar{v}_i})}$ be a free vector passing through a point $\Vect{p_i(\Vect{\bar{p}_i})}$ on a rigid body $i$ and similarly we define terms for another rigid body $j$. Then a revolute joint connection between rigid body $i$ and $j$ imposes the kinematic constraint $\Vect{C^{rev}_{ij}} =  [(\Vect{p_j(\Vect{\bar{p}_j})} - \Vect{p_i(\Vect{\bar{p}_i})})^T \quad (\Vect{v_j(\Vect{\bar{v}_j})} - \Vect{v_i(\Vect{\bar{v}_i})})^T ]^T $. $\Vect{C^{rev}_{ij}}$ consists of $6$ scalar constraints, first and second $3$ scalar constraints measure distance between each component of two points and two vectors, respectively. When these $6$ scalar constraints have zero value, they define the rotation axis of the revolute joint passing through coincident points of two rigid bodies. It should be noted that there are effectively only $5$ independent scalar constraints which allow $1$-DOF for a revolute joint connection.}
{A spherical joint connection imposes the kinematic constraint $\Vect{C^{sph}_{ij}} =  (\Vect{p_j(\Vect{\bar{p}_j})} - \Vect{p_i(\Vect{\bar{p}_i})})$. $\Vect{C^{rev}_{ij}}$ consists of $3$ scalar constraints. When these three scalar constraints have zero value, a spherical joint connection is created with a coincident center of rotation. There are effectively $3$ independent scalar constraints which allow $3$-DOF between the rigid bodies $i$ and $j$ for a spherical joint connection. For more details, please refer to~\cite{coros_computational_2013, thomaszewski_computational_2014}.}

\subsubsection{Omnidirectional mobile bases}
{Each mobile base $k = 0,1,2 \cdots n_{m}-1$ is also counted as a rigid body with state $\Vect{s^m_k}=[\gamma^m_k \quad \beta^m_k \quad \alpha^m_k \quad {x^m_k} \quad {y^m_k} \quad {z^m_k}]^T$. Let $\Vect{\bar{vn}^m_k}$  be a unit vector attached to planar mobile base, which is perpendicular to its plane.
For a world reference frame ($\Vect{O_g}$, $\mathbf{x_g}$, $\mathbf{y_g}$, $\mathbf{z_g}$) $\mathbf{x_g}$, $\mathbf{y_g}$ and $\mathbf{z_g}$ are three unit vectors and $\Vect{O_g} = [0 \quad 0 \quad 0]^T$. For each mobile base, we add kinematic constraints $\Vect{C^{planar}_k} = [{z^m_k} \quad (\Vect{vn^m_k(\bar{vn}^m_k)} -  \mathbf{z_g})^T]^T$ which correspond to the planar constraint on the mobile robot. This planar constraint restricts translation along $\mathbf{z_g}$ and rotation about $\mathbf{x_g}$ and $\mathbf{y_g}$. We further add motor constraints assuming motorized prismatic actuators along $\mathbf{x_g}$, ${C^{mx}_k}={x^m_k} - \Vect{u}[3\cdot k]$, and along $\mathbf{y_g}$, ${C^{my}_k}={y^m_k} - \Vect{u}[3\cdot k+ 1]$. Let $\Vect{\bar{vp}^m_k} = [1 \quad 0 \quad 0]^T$ be a unit vector in the plane of the mobile robot and $\theta = \Vect{u}[3\cdot k +2]$ be angle between $\Vect{vp^m_k(\bar{vp}^m_k)}$ and $\mathbf{x_g}$ about axis $\mathbf{z_g}$. We further add motor constraints assuming motorized rotary actuator about axis $\mathbf{z_g}$, $\Vect{C^{mz}_k}=\Vect{R_{\gamma^m_k}} \cdot \Vect{R_{\beta^m_k}} \cdot \Vect{R_{\alpha^m_k}}\cdot (\Vect{R_{\theta}}\cdot \Vect{\bar{vp}^m_k}) - \mathbf{x_g}$. For the \textit{reduced actuation scheme}, $\theta$ is free and can be fixed to an arbitrary number.}

{The vector of constraints $\Vect{C}$ includes all these constraints output by each passive joint (revolute, prismatic, spherical, universal) between the two rigid bodies, motor constraints obtained by fixing the value of actuator in the motorized joint between the two rigid bodies. It should be noted that the vector of constraints $\Vect{C}$ is both a function of $\Vect{s}$ and $\Vect{u}$.
}

%Each mobile base is counted as a rigid body with $6$-DOF with additional constraints based on its type. In the case of omnidirectional bases, mobile agents have three constraints (planar constraint) on its motion reducing its DOF to $3$. Following presents the definitions of different variables used in the paper:

\section{Optimization framework for simulation and kinematic control of CCMA system}
\label{sec:model_opti}
\subsection{Simulation of the CCMA system}
For simulation of the CCMA system, we solve an optimization problem where we minimize an energy $E(\mathbf{s}, \mathbf{u})$ which is a function of state $\mathbf{s}$ and control variables in $\mathbf{u}$. The vector $\Vect{C}$ contains all the kinematic constraints including those imposed by the passive kinematic chain and the actuation of multiple mobile agents.
\begin{equation}
\begin{aligned}
\Vect{\hat{s}} = & \underset{\mathbf{s(u)}}{\text{minimize}}
& & E(\mathbf{s}, \mathbf{u}) = \frac{1}{2}\mathbf{C(\mathbf{s}, \mathbf{u})}^{T}\mathbf{C(\mathbf{s}, \mathbf{u})} \\
\end{aligned}
\label{eqn:opt1}
\end{equation}
This allows us to simulate the CCMA system and solve the forward kinematics problem which is to find the complete state $\Vect{\hat{s}}$, including the designated end-effector state, $\mathbf{X_{EE}}$,  when the control vector $\mathbf{u}$ is given as input. We calculate analytically the first and second order derivatives $\frac{dE}{d\Vect{s}}, \frac{d^2E}{d\Vect{s}^2}$ and utilize standard Newton Raphson method to solve this minimization problem in iterative manner as follows:
\begin{equation*}
\Vect{s_{i+1}} = \Vect{s_i} - \evalat[\Bigg]{\Bigg(\frac{d^2E}{d\Vect{s}^2}\Bigg)^{-1}}{\Vect{s} = \Vect{s_i}} \cdot \evalat[\Bigg]{\frac{dE}{d\Vect{s}}}{\Vect{s} = \Vect{s_i}}
\end{equation*}
Since, $\Vect{C}$ and the resulting constraint energy term $E$ are abstracted to the level of type of joints or what type of actuators are used, the analytical derivation of the $\frac{dE}{d\Vect{s}}, \frac{d^2E}{d\Vect{s}^2}$ is system independent.
%of the choice of design, morphology of the passive kinematic chain or the number, actuation scheme of the mobile agents. 
This allows us to plug and play different design, topologies of the passive kinematic chain and different actuation schemes, number of mobile agents, in the CCMA system. 

\subsection{Kinematic control of the CCMA system}
{In order to solve the kinematic control problem, which is to find $\mathbf{u}$ for desired end-effector state $\mathbf{X^{*}_{EE}}$, we solve the optimization problem in Eqn.~\ref{eqn:opt3}. The minimization in Eqn.~\ref{eqn:opt1} only ensures that the gradient of energy $E(\mathbf{s}, \mathbf{u})$, $\Vect{G}$, is zero upon convergence and not the energy $E(\mathbf{s}, \mathbf{u})$ itself.  This would mean that CCMA system does not assemble properly or that the kinematic constraints are not satisfied. 
%This is not acceptable for simulation and control of the CCMA system, as in each state $\Vect{s}$ the CCMA system should satisfy all the kinematic constraints. 
We solve this problem by adding the residual constraint energy $E_r = E(\mathbf{\hat{s}}, \mathbf{u})$ in the objective function $\mathcal{O}$, where $0<\lambda<1$.}
%, in Eqn.~\ref{eqn:opt2}. $E_r  is the residual constraint energy $E_r$
\begin{multline}
 \underset{\mathbf{u}}{\text{minimize}} \quad
 \mathcal{O}(\mathbf{s(u)}, \mathbf{u}) =\\ \frac{1}{2}(\mathbf{X_{EE}-X^{*}_{EE}})^{T}(\mathbf{X_{EE}-X^{*}_{EE}})
+ \lambda \cdot E(\mathbf{\hat{s}}, \mathbf{u})
\label{eqn:opt3}
\end{multline}
{We also need to calculate the derivatives of the objective function $\mathcal{O}$, as in Eqn.~\ref{eqn:derivOpti2E}, with respect to the control variables, in order to use gradient based optimization techniques.}
\begin{equation}
\frac{d\mathcal{O}}{d\mathbf{u}} = \frac{\partial \mathcal{O}}{\partial \mathbf{X_{EE}}} \cdot
\frac{\partial  \mathbf{X_{EE}}}{\partial \mathbf{s}} \cdot \frac{d\mathbf{s}}{d\mathbf{u}} + \lambda \cdot \frac{dE_r}{d\Vect{u}}
\label{eqn:derivOpti2E}
\end{equation}
%\begin{equation}
%\begin{aligned}
% \underset{\mathbf{u}}{\text{minimize}} \quad
% &\mathcal{O}(\mathbf{s(u)}, \mathbf{u})\\ = &\frac{1}{2}(\mathbf{X_{EE}-X^{*}_{EE}})^{T}(\mathbf{X_{EE}-X^{*}_{EE}}) \\
%\end{aligned}
%\label{eqn:opt2}
%\end{equation}
%\begin{equation}
%\frac{d\mathcal{O}}{d\mathbf{u}} = \frac{\partial \mathcal{O}}{\partial \mathbf{X_{EE}}} \cdot
%\frac{\partial  \mathbf{X_{EE}}}{\partial \mathbf{s}} \cdot \frac{d\mathbf{s}}{d\mathbf{u}}
%\label{eqn:derivOpt2}
%\end{equation}
The analytical expression for $\frac{\partial \mathcal{O}}{\partial \mathbf{X_{EE}}} \cdot\frac{\partial  \mathbf{X_{EE}}}{\partial \mathbf{s}}$ is easily obtained. 
However, the expression for $\frac{d\mathbf{s}}{d\mathbf{u}}$ is not generally analytically available and use of finite differences to compute it will require to minimize the constraint energy function $E(\mathbf{s}, \mathbf{u})$ $2$k times to high degree of accuracy, where k is the size of the control vector  $\mathbf{u}$. This is computationally too demanding for a real time simulation and control tool. We solve this problem by doing the sensitivity analysis over the gradient $\frac{dE}{d\Vect{\Vect{s}}}$ of the constraint energy function  $E(\mathbf{s}, \mathbf{u})$ rather than over the constraint vector $\mathbf{C}$ itself as done in the paper~\cite{coros_computational_2013}. This is done because constraints can not be assumed to be satisfied during the intermediate iterations of the optimization of the objective function $\Vect{\mathcal{O}}$. On the other hand, the gradient $\frac{dE}{d\Vect{s}}$ of the constraint energy function is always equal to zero, when the optimization in Eqn.~\ref{eqn:opt1} converges. To express it more clearly for every control vector $\mathbf{u}$, one can find a suitable $\mathbf{s}$ such that gradient of the constraint energy function $E(\mathbf{s}, \mathbf{u})$ is zero. Thus we have the following identity:
\begin{equation}
\mathbf{G} = \frac{dE}{d\mathbf{s}} = 0 \quad\forall \Vect{u} \implies \quad \quad\quad \quad \frac{d\mathbf{G}}{d\mathbf{u}} = 0 \quad \quad 
\label{eqn:senA1}
\end{equation}
Doing the sensitivity analysis over this gradient $\Vect{G}$ leads to the following equation:
\begin{equation}
 \frac{\partial \mathbf{G}}{\partial \mathbf{u}} + \frac{\partial \mathbf{G}}{\partial \mathbf{s}} \cdot\frac{d\mathbf{s}}{d\mathbf{u}} = 0 \quad \quad \quad
\frac{d\mathbf{s}}{d\mathbf{u}} = -\Bigg(\frac{\partial \mathbf{G}}{\partial \mathbf{s}}\Bigg)^{-1}\cdot\frac{\partial \mathbf{G}}{\partial \mathbf{u}}
\label{eqn:senA2}
\end{equation}
$\frac{\partial \mathbf{G}}{\partial \mathbf{s}} = \frac{d^2E}{d\Vect{s}^2}$ is the Hessian of the constraint energy function $E(\mathbf{s}, \mathbf{u})$, which we calculate analytically. $\frac{\partial \mathbf{G}}{\partial \mathbf{u}}$ is the sensitivity of the gradient $\mathbf{G}$ with respect to control vector $\Vect{u}$, while state $\Vect{s}$ is kept constant. 
%The calculation of this term using finite differences is not computationally demanding, as the state $\mathbf{s}$ is assumed to be constant and the constraint energy function $E$ does not need to be minimized. 
We analytically calculate this term $\frac{\partial \mathbf{G}}{\partial \mathbf{u}}$ as well. With the analytical expressions for $\frac{\partial \mathbf{G}}{\partial \mathbf{s}}$ and $\frac{\partial \mathbf{G}}{\partial \mathbf{u}}$, finally we can compute $\frac{d\mathbf{s}}{d\mathbf{u}}$. 

$\evalat[\Big]{\frac{\partial E}{\partial \Vect{s}}}{\Vect{s} = \Vect{\hat{s}}}=0$ because $\Vect{s} = \Vect{\hat{s}}$ is solution to Eqn.~\ref{eqn:opt1}.
Eqn.~\ref{eqn:derivOpti2E} requires calculation of an additional term $\frac{dE_r}{d\Vect{u}}$ which is done as follows: 
\begin{equation}
\frac{dE_r}{d\Vect{u}} = \evalat[\Bigg]{\frac{\partial E}{\partial \Vect{u}}}{\Vect{s} = \Vect{\hat{s}}} + \evalat[\Bigg]{\frac{\partial E}{\partial \Vect{s}}}{\Vect{s} = \Vect{\hat{s}}} \cdot \frac{d\Vect{s}}{d\Vect{u}} = \evalat[\Bigg]{\frac{\partial E}{\partial \Vect{u}}}{\Vect{s} = \Vect{\hat{s}}}
\label{eqn:epu}
\end{equation}
We calculate this term ${\frac{\partial E_r}{\partial \Vect{u}}}$, in the Eqn.~\ref{eqn:epu}, analytically. 

{Substituting expression for $\frac{d\mathbf{s}}{d\mathbf{u}}$ and $\frac{dE_r}{d\Vect{u}}$  in Eqn.~\ref{eqn:derivOpti2E}, we can calculate the gradient $\frac{dO}{d\mathbf{u}}$ of the objective function $\Vect{\mathcal{O}}$.} This gradient can then be used to update the control vector $\mathbf{u}$ in the current step of the optimization problem as follows:
\begin{equation*}
\mathbf{u_{i+1}}=\Vect{u_{i}} - \hat{\Vect{H}}^{-1}\cdot
\evalat[\Bigg]{\frac{d\mathcal{O}}{d\mathbf{u}}}{\Vect{u} = \Vect{u_i}}
\end{equation*}
For exact Newton method, we had need anlytical expression for the derivative of the Hessian of  $E(\mathbf{s}, \mathbf{u})$ with respect to state $\mathbf{s}$, which we don't calculate. Instead, we approximate $\hat{\Vect{H}}^{-1}$ using the BFGS quasi-Newton method.

\subsection{Results on convergence and computational effort}
\begin{figure}[!t] 
\vspace{7pt}
  \centering
    \includegraphics[width=1.0\columnwidth]{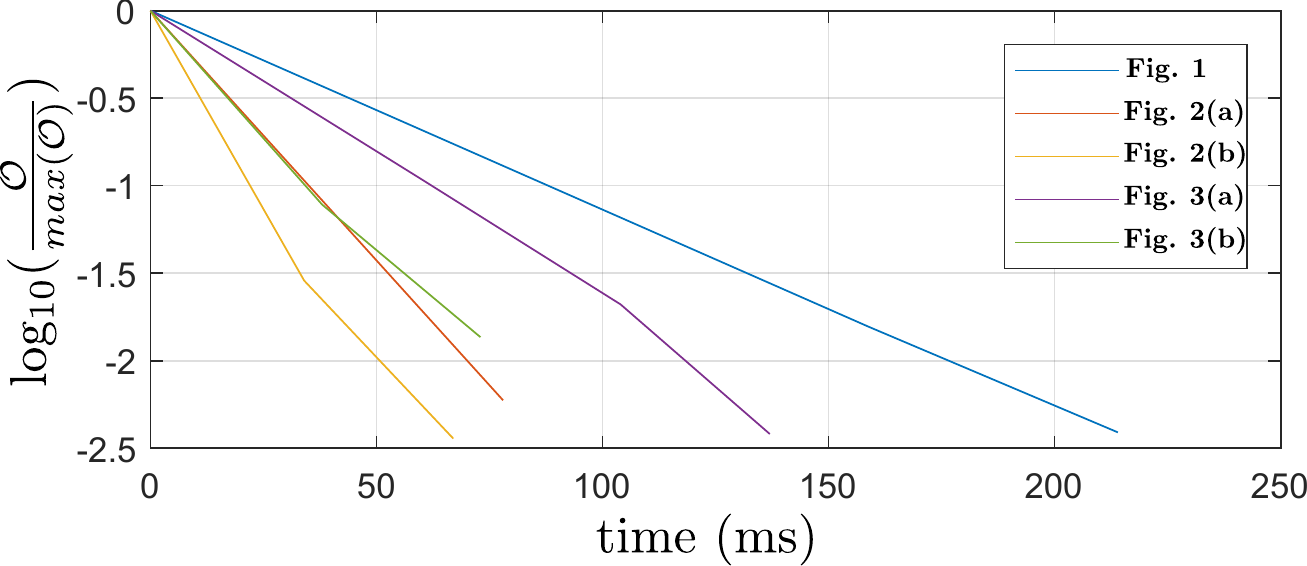}
    \caption{Convergence plots with respect to time for CCMA system examples in Fig.~\ref{fig:concepCCMA},~\ref{fig:4DOFExamples}(a),~\ref{fig:4DOFExamples}(b),~\ref{fig:6DOFExamples}(a) and~\ref{fig:6DOFExamples}(b).}
    \label{fig:convergencePlots}
\end{figure}

While it is possible to directly calculate the final control vector $\mathbf{u}$ for large changes in $\mathbf{X_{EE}}$. It is advisable to do small steps both in end-effector position and orientation until we reach the desired $\mathbf{X^{*}_{EE}}$, especially for actual hardware experiments. This leads to smoother motion of the CCMA system, in particular during end-effector rotations. Fig.~\ref{fig:convergencePlots} presents the computational effort, in time, which is required to achieve a small combined perturbation in the $\mathbf{X_{EE}}$, $10$~mm step size for translation and $0.005$~radians step size for rotation. 
The convergence and computational effort is presented for different CCMA system examples across different designs, morphologies of the passive kinematic chain and different number, actuation schemes of the mobile agents.
It can be observed that the computational effort is smallest for the $4$-DOF CCMA system with $3$ mobile agents and highest for $6$-DOF CCMA system with $6$-mobile agents. It should be noted that, for each plot, a combined perturbation in $\mathbf{X_{EE}}$ was solved.

\section{Experimental results}
\label{sec:results}
In this section, we present the fabricated prototypes for the $4$-DOF CCMA system with two different actuation schemes.
%, corresponding to Fig.~\ref{fig:4DOFExamples}. 
Furthermore, we demonstrate the kinematic control of fabricated prototypes using the optimization framework developed in previous section and describe experimental results.

In order to determine and track the position and orientation of the mobile bases, with respect to the world reference frame, a motion capture system comprising of $10$ OptiTrack Prime $13$ cameras
%, as shown in Fig.~\ref{fig:mocap}, 
was used. 
%\begin{figure}[!b] 
%  \centering
%    \includegraphics[width=0.9\columnwidth]{mocap}
%    \caption{Optitrack motion capture system with four stands having $10$ cameras on the four corners of the room}
%    \label{fig:mocap}
%\end{figure}
\begin{figure}[!t] 
\vspace{7pt}
  \centering
    \includegraphics[width=1.0\columnwidth]{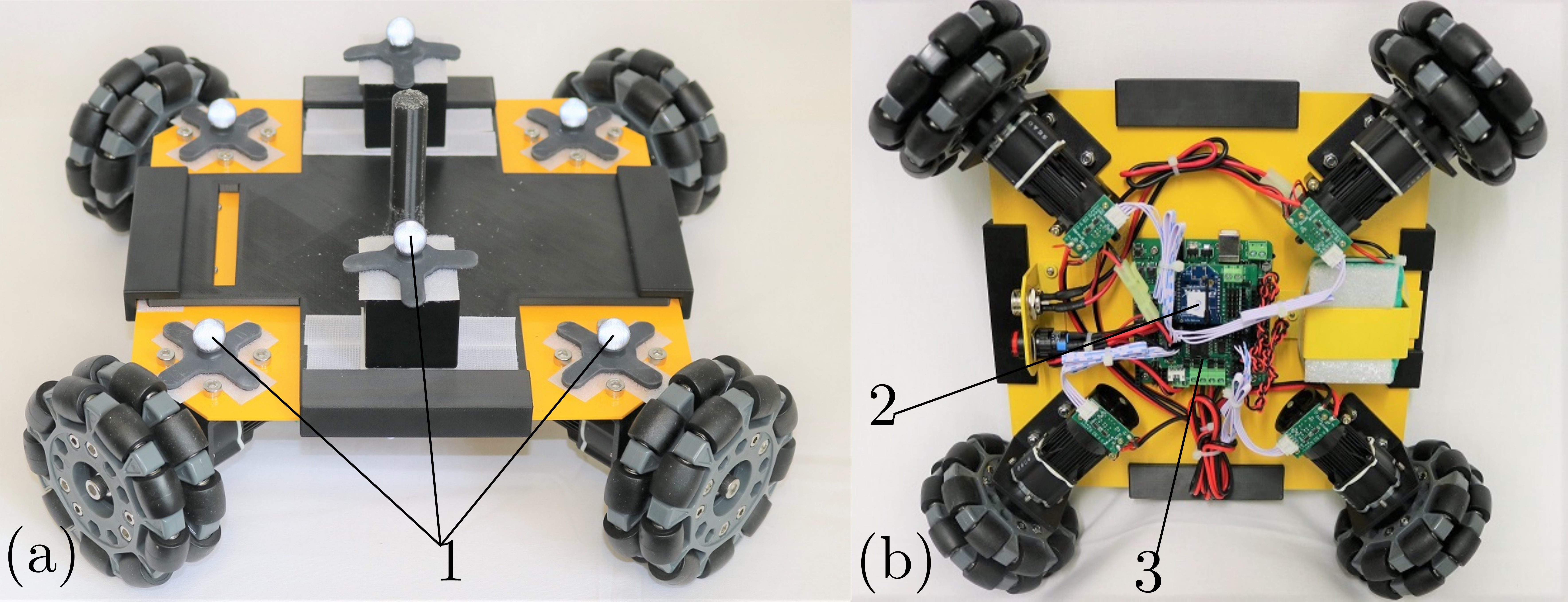}
    \caption{(a) Top view of the omnidirectional wheeled mobile base with 1) optical markers (b) Bottom view with a 2) wireless transceiver XBee and a 3) micro-controller}
    \label{fig:omniMobileBase}
\end{figure} 
Fig.~\ref{fig:omniMobileBase} shows the omnidirectional base used for proof of concept demonstrators in the paper. Each mobile base has $6$ optical markers (Fig.~\ref{fig:omniMobileBase}(a)) which can be tracked with the motion capture system. Motion capture system sends the feedback for the actutal mobile base position and orientation in the world reference frame to the computer running the optimization. In order to wirelessly transmit the control signals to the mobile bases, for the control of the CCMA demonstrators, we use wireless transreceivers XBees from Digi International. Each mobile base is equipped on its back with an XBee (Fig.~\ref{fig:omniMobileBase}(b)). The computer running the optimization sends the control commands to a micro-controller via serial communication. This micro-controller has a transreceiver XBee, which acts as the co-ordinator. The co-ordinator XBee broadcasts the control signals wirelessly to the respective XBees, in each mobile base. 

The optimization framework, as described in Sec.~\ref{sec:model_opti}, calculates the states of the mobile bases for desired end-effector motions. The state of each mobile base is then converted to the corresponding wheel speeds based on the difference between the set and the tracked state of the mobile base, using a PI (proportional integral) controller. Please refer to the standard mobile robot kinematics for swedish wheels based omnidirectional mobile robot in this work~\cite{siegwart_introduction_2004}. It should be noted that the closed loop feedback PI controller is used to track the states of the mobile bases only. For the experimental results described hereafter, end-effector motion of the CCMA system is still obtained via feed-forwarding the computed mobile base states in an open loop. 

\begin{figure}[!t] 
\vspace{7pt}
  \centering
    \includegraphics[width=0.85\columnwidth]{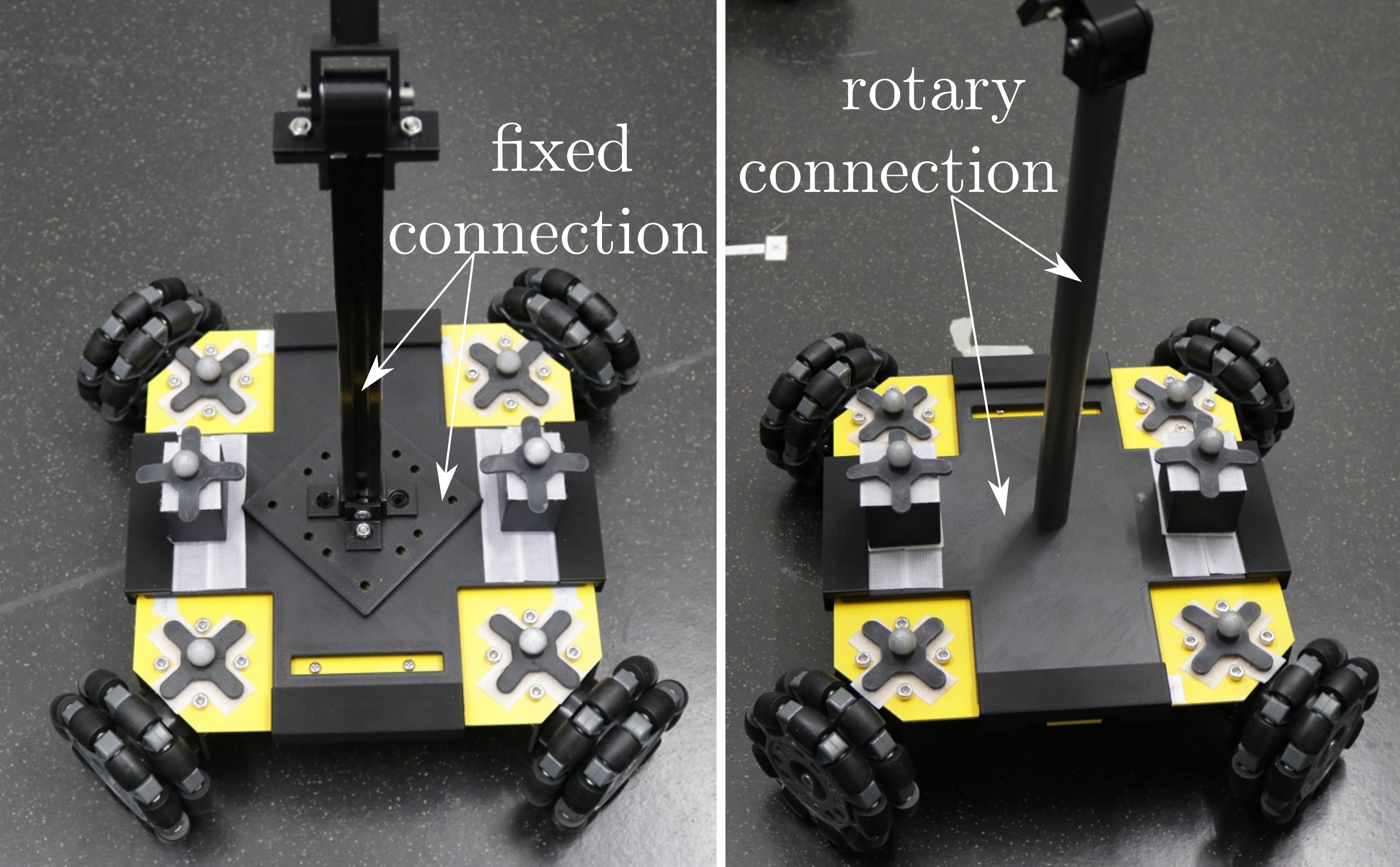}
    \caption{(a) Mobile base has a fixed bolted connection with the link adjacent to it. This is the fabrication strategy for complete actuation schemes. (b) Mobile base has a rotary connection with the link adjacent to it. This is the fabrication strategy for \textit{reduced actuation schemes}.}
    \label{fig:mobileBaseActuationScheme}
\end{figure}

\subsection{Reduced actuation scheme}
\begin{figure}[!t] 
%\vspace{7pt}
  \centering
    \includegraphics[width=0.85\columnwidth]{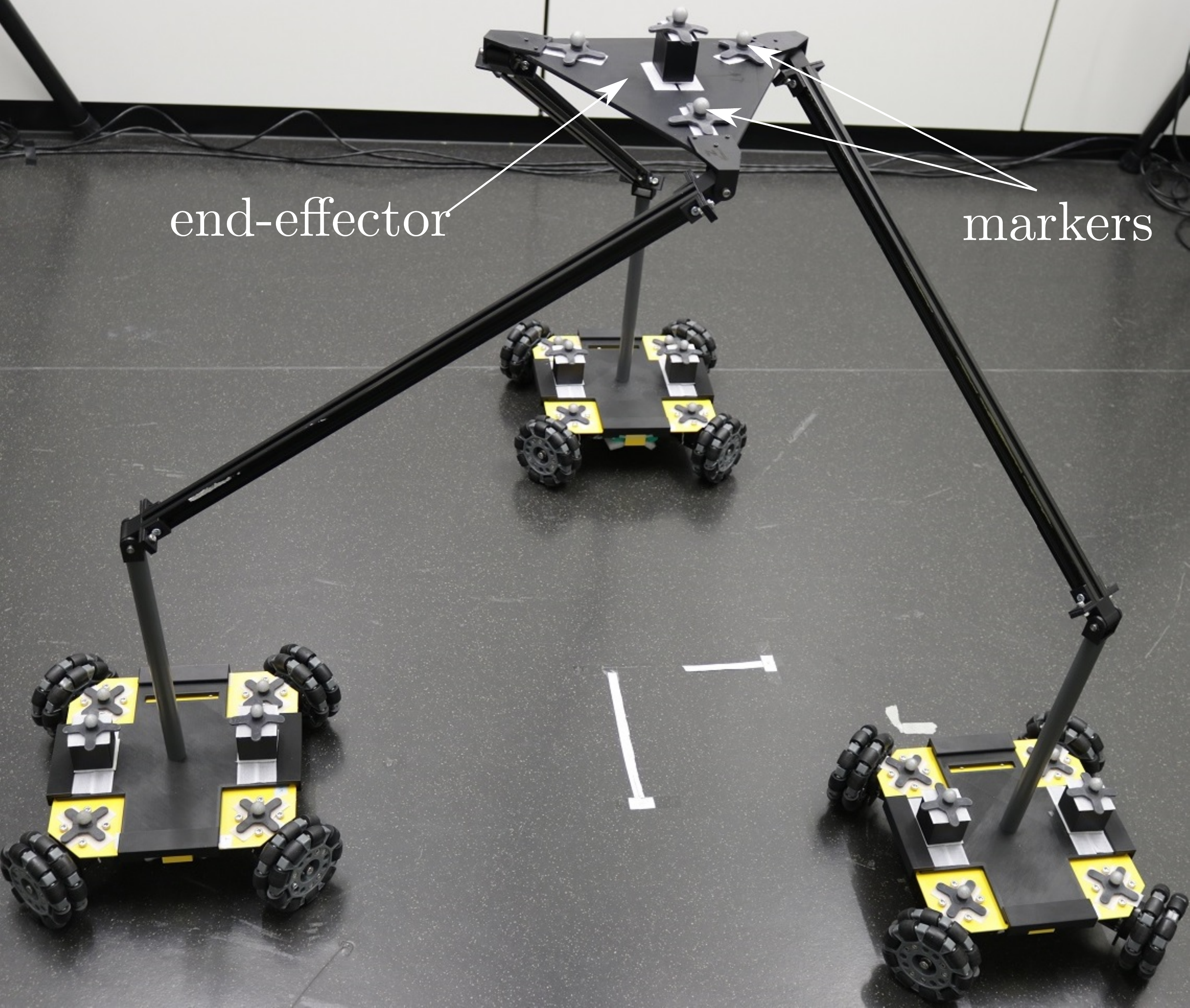}
    \caption{Fabricated prototype of $4$-DOF CCMA system example from Fig.~\ref{fig:4DOFExamples}(a) with optical markers.}
    %in the end-effector for its tracking with the mocap system.}
    \label{fig:4DOFprotos}
\end{figure}
\begin{figure}[!t] 
  \centering
    \includegraphics[width=0.9\columnwidth]{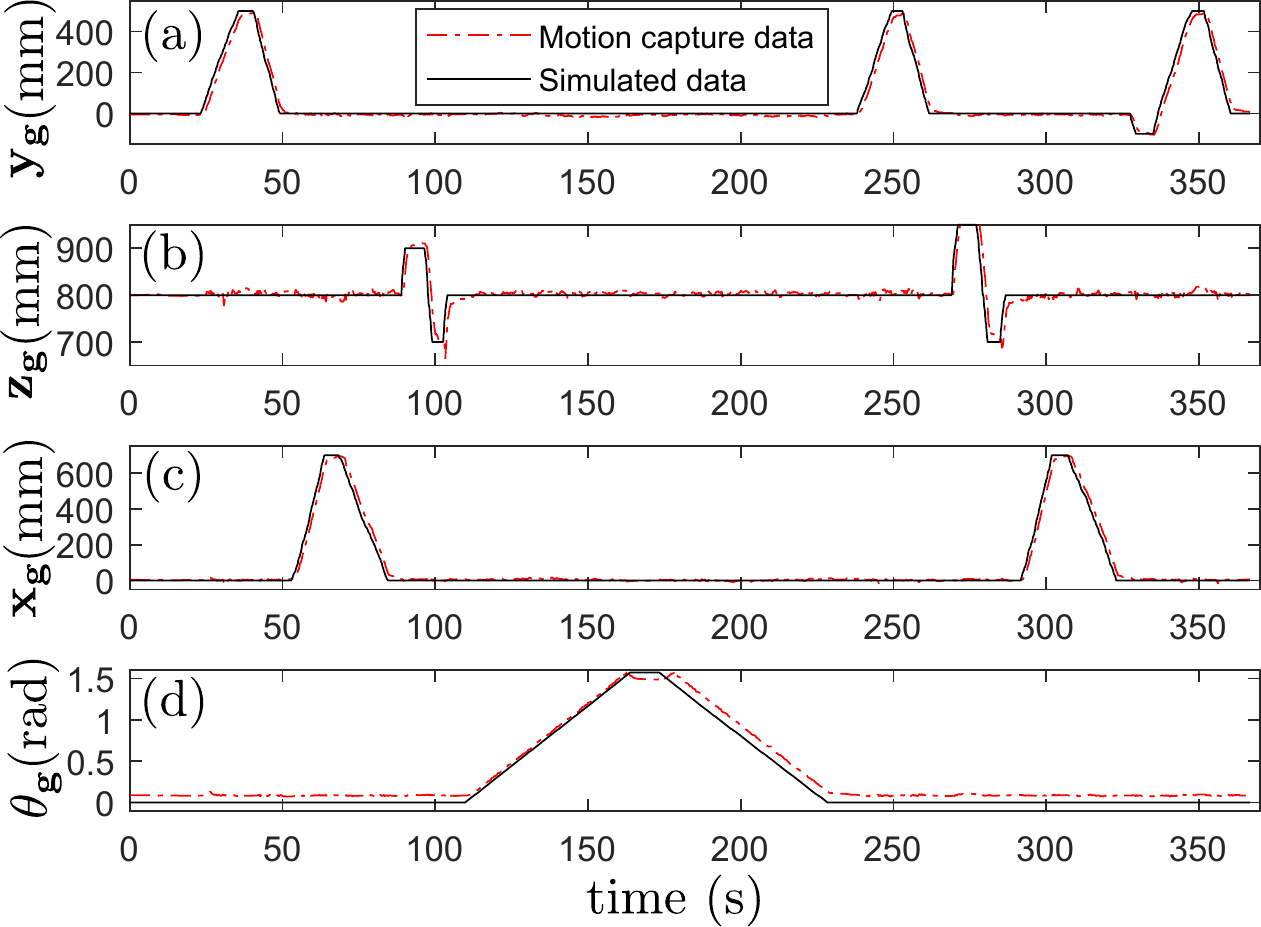}
    \caption{The tracked and simulated motion of the end-effector with reduced actuation scheme (a) translation along $\Vect{y_g}$ (b) along $\Vect{z_g}$ (c) along $\Vect{x_g}$ (d) rotation about $\Vect{z_g}$.}
    \label{fig:data1xEETracking}
\end{figure}
\begin{figure}[!t] 
\vspace{7pt}
  \centering
    \includegraphics[width=0.9\columnwidth]{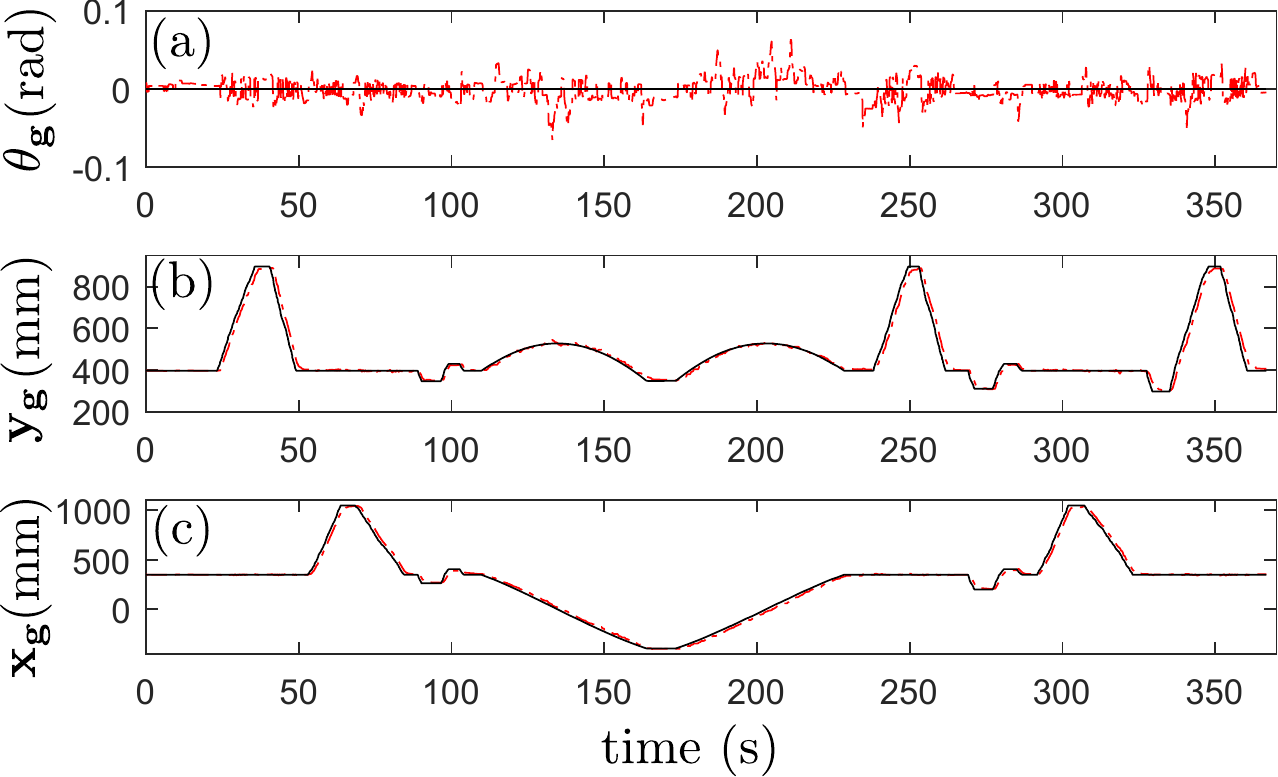}
    \caption{The tracked and simulated motion of one of the mobile bases generated by the optimization framework for the end-effector motion in Fig.~\ref{fig:data1xEETracking}. (a) rotation about $\Vect{z_g}$ (b) translation along $\Vect{y_g}$ (c) translation along $\Vect{x_g}$.}
    \label{fig:data1mb2TrackingTime}
\end{figure}
\begin{figure}[!t] 
\vspace{7pt}
  \centering
    \includegraphics[width=0.9\columnwidth]{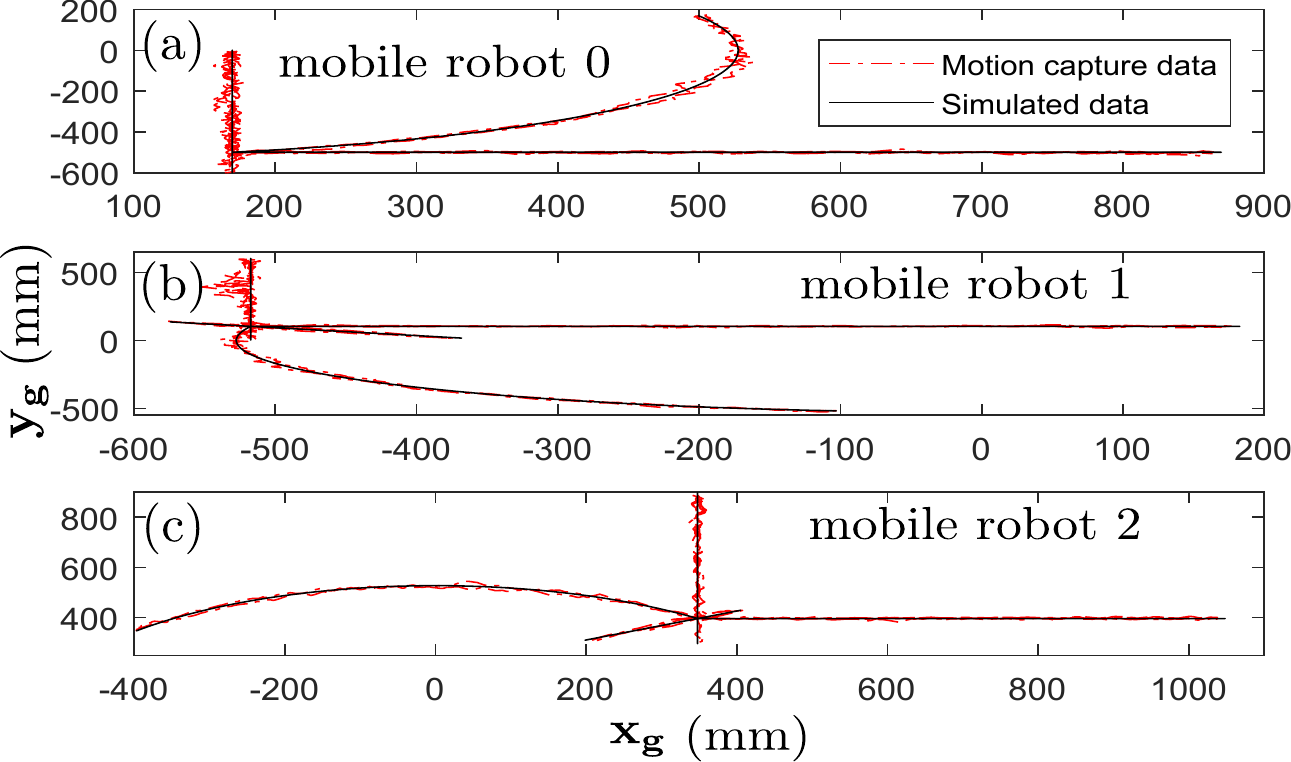}
    \caption{Motion of the three mobile bases, in the ground plane ($\Vect{x_g} - \Vect{y_g}$), generated by the optimization framework for the desired end-effector motion in Fig.~\ref{fig:data1xEETracking}.}
    \label{fig:data1mbTrackingXY}
\end{figure}
\begin{figure}[!t] 
  \centering
    \includegraphics[width=0.9\columnwidth]{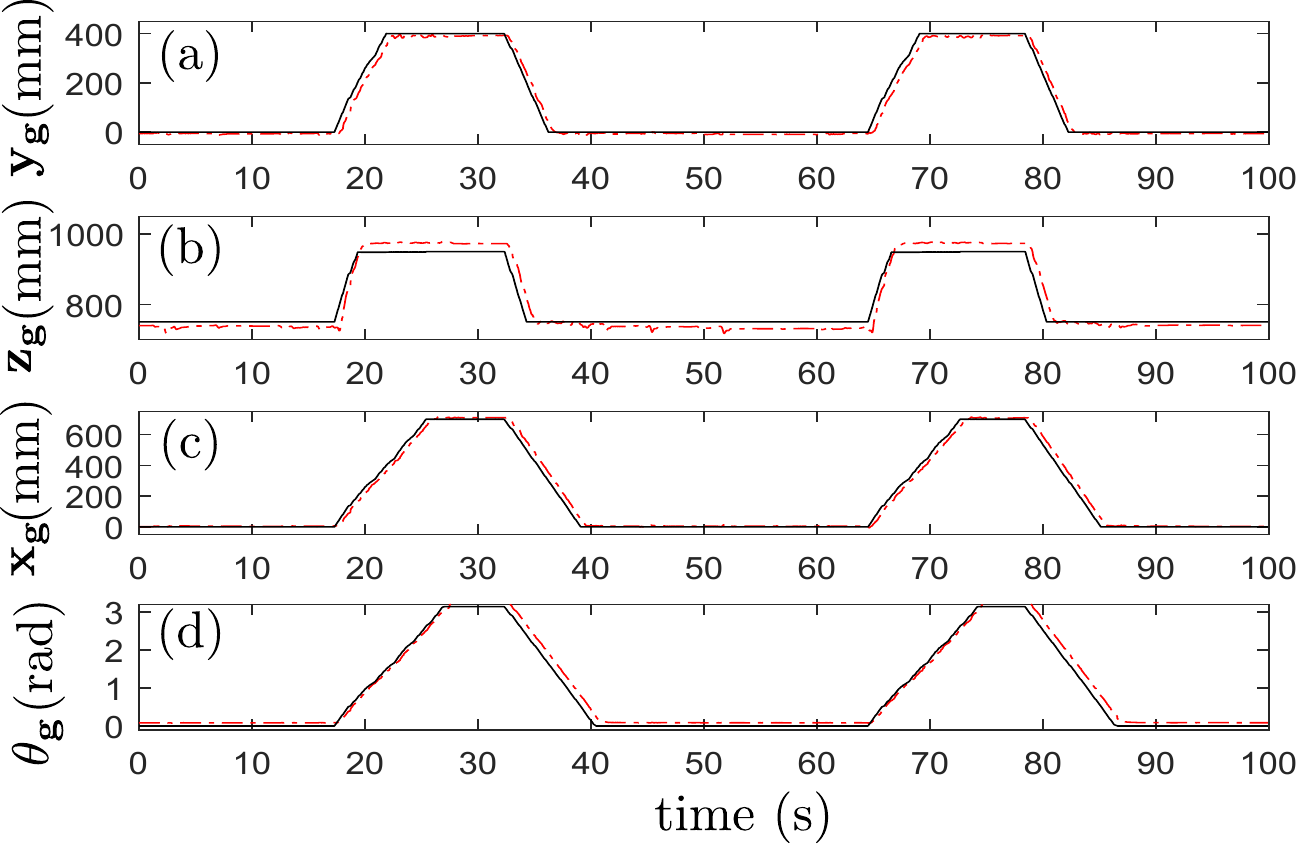}
    \caption{The tracked and simulated motion of the end-effector along each DOF simultaneously (prototype with reduced actuation scheme) (a) translation along $\Vect{y_g}$ (b) along $\Vect{z_g}$ (c) along $\Vect{x_g}$ (d) rotation about $\Vect{z_g}$.}
    \label{fig:data2xEETracking}
\end{figure}
Fig.~\ref{fig:4DOFprotos} shows the fabricated prototype of the $4$-DOF CCMA system. It can translate along $\Vect{x_g}, \Vect{y_g}, \Vect{z_g}$ and rotate about $\Vect{z_g}$. Between the mobile base and the adjacent link, there is a rotary connection obtained by use of two concentric cylinders (Fig.~\ref{fig:mobileBaseActuationScheme}(b)). Therefore, this particular prototype utilizes the \textit{reduced actuation scheme} and $2$-DOF in translation of the mobile bases (size of control vector $\Vect{u}$ is $6$) to manipulate the $4$-DOF end-effector. For experiments with this prototype, the orientation of the mobile base is kept constant. Fig.~\ref{fig:data1xEETracking} shows the sequential movement along each DOF of the end-effector with respect to time. Fig.~\ref{fig:data1mb2TrackingTime} and Fig.~\ref{fig:data1mbTrackingXY} shows the corresponding motion of the mobile bases with respect to time and the ground plane ($\Vect{x_g} - \Vect{y_g}$), respectively. RMSE (Root Mean Square Error) were $18$, $12$, $17$~mm and $0.09$~rad for top to bottom plots, respectively, for the end-effector motion in Fig.~\ref{fig:data1xEETracking}.

\begin{figure}[!t] 
\vspace{7pt}
  \centering
    \includegraphics[width=0.9\columnwidth]{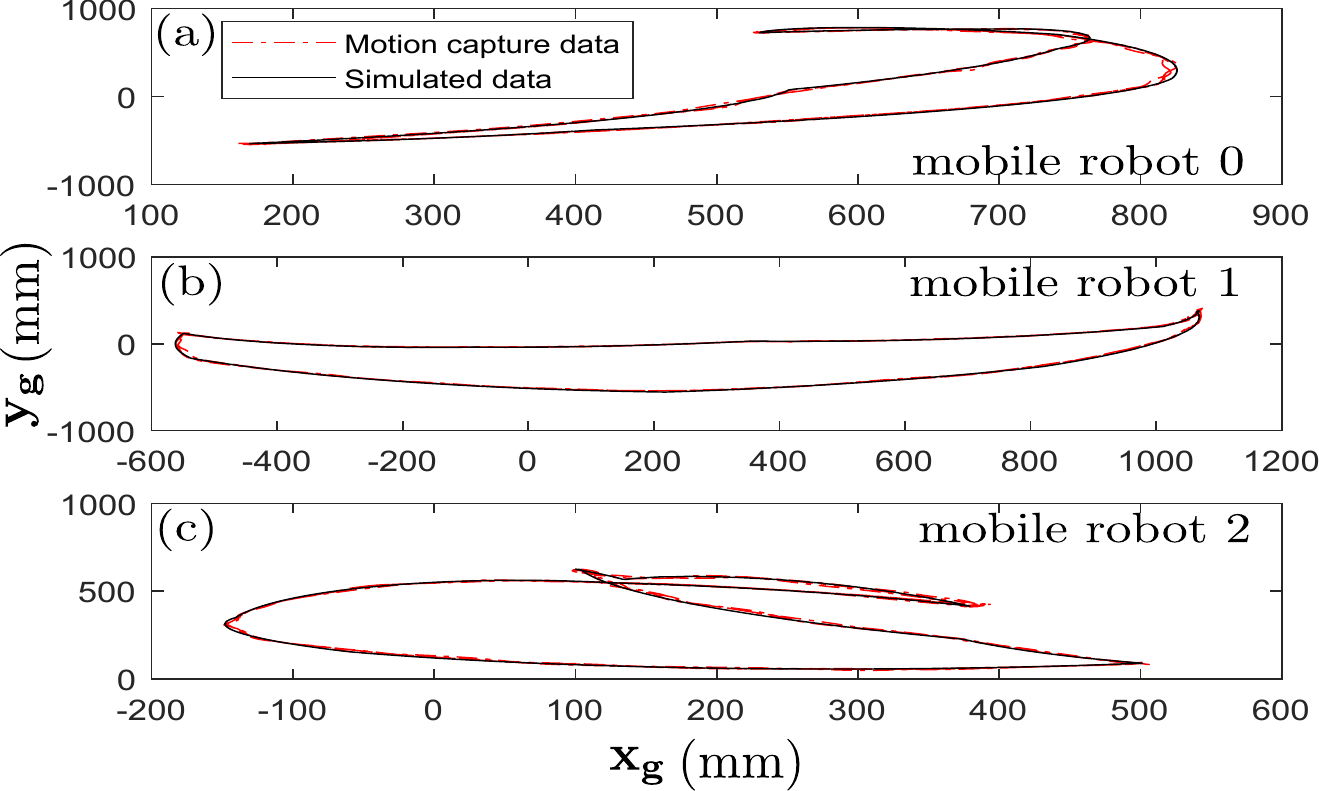}
    \caption{Motion of the three mobile bases, in the ground plane ($\Vect{x_g} - \Vect{y_g}$), generated by the optimization framework for the desired end-effector motion in Fig.~\ref{fig:data2xEETracking}.}
    \label{fig:data2mbTrackingXY}
\end{figure}

Fig.~\ref{fig:data2xEETracking} shows the combined simultaneous movement, along each DOF of the end-effector, with respect to time. Fig.~\ref{fig:data2mbTrackingXY} show the corresponding motion of the mobile bases with respect to the ground plane ($\Vect{x_g} - \Vect{y_g}$). RMSE were $18$, $22$, $24$~mm and $0.13$~rad for top to bottom plots, respectively, for the end-effector motion in Fig.~\ref{fig:data2xEETracking}.
 
\subsection{Complete actuation scheme}
We also fabricate a different $4$-DOF prototype with fixed connection (Fig.~\ref{fig:mobileBaseActuationScheme}(a)), which utilizes complete actuation scheme and full $3$-DOF of the mobile bases (size of control vector $\Vect{u}$ is $9$) to manipulate the $4$-DOF end-effector. 
\begin{figure}[!t] 
  \centering
    \includegraphics[width=0.9\columnwidth]{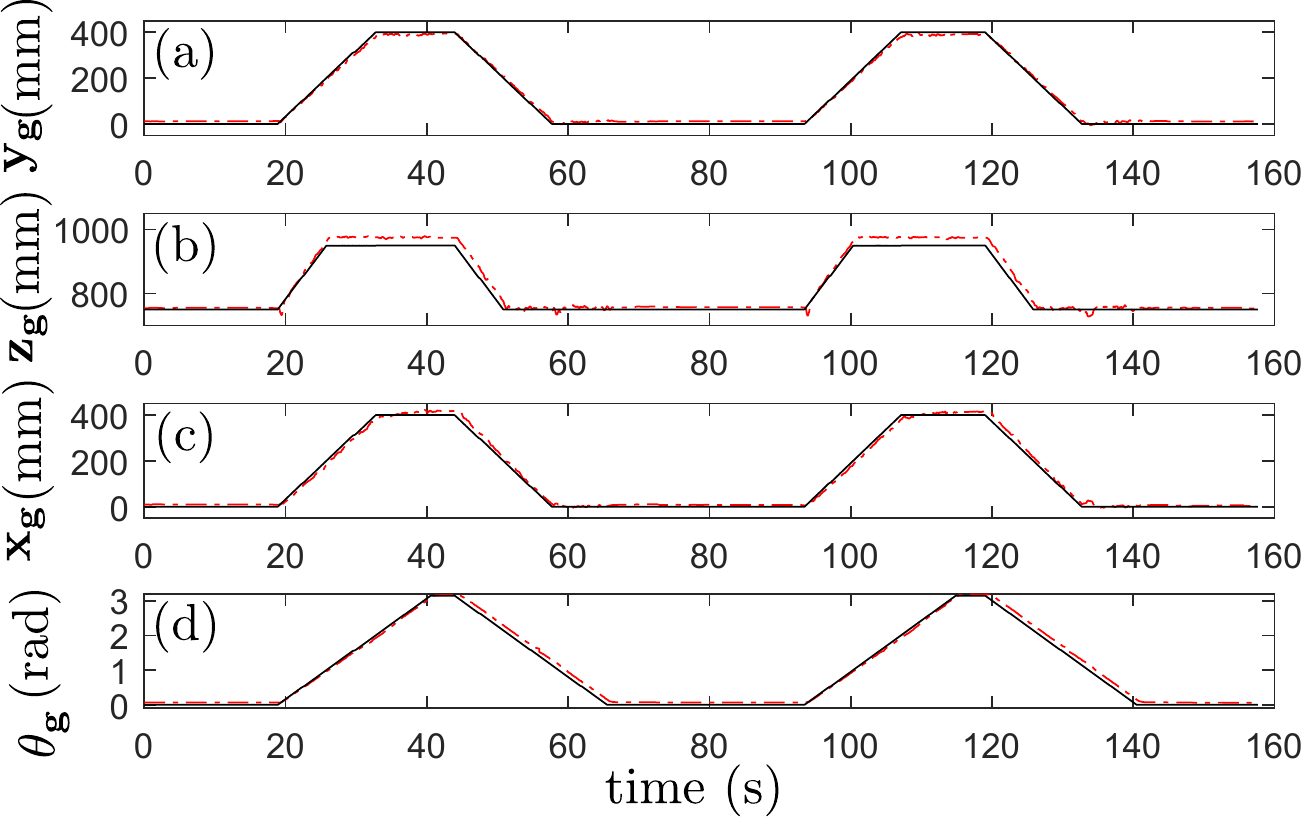}
    \caption{The tracked and simulated motion of the end-effector along each DOF simultaneously (prototype with complete actuation scheme) (a) translation along $\Vect{y_g}$ (b) along $\Vect{z_g}$ (c) along $\Vect{x_g}$ (d) rotation about $\Vect{z_g}$.}
    \label{fig:data3xEETracking}
\end{figure}
\begin{figure}[!t] 
\vspace{7pt}
  \centering
    \includegraphics[width=0.9\columnwidth]{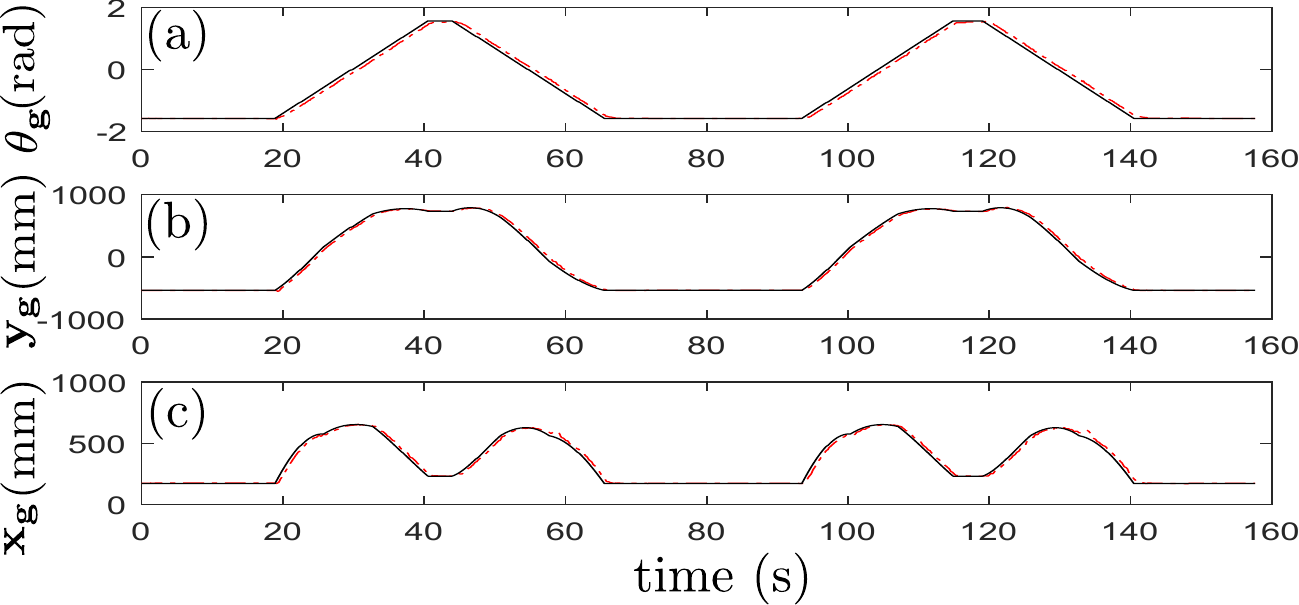}
    \caption{The tracked and simulated motion of one of the mobile bases generated by the optimization framework for the end-effector motion in Fig.~\ref{fig:data3xEETracking} (a) rotation about the $\Vect{z_g}$ (b) translation along $\Vect{y_g}$ (c) translation along $\Vect{x_g}$.}
    \label{fig:data3mb0TrackingTime}
\end{figure}
\begin{figure}[!t] 
  \centering
    \includegraphics[width=0.9\columnwidth]{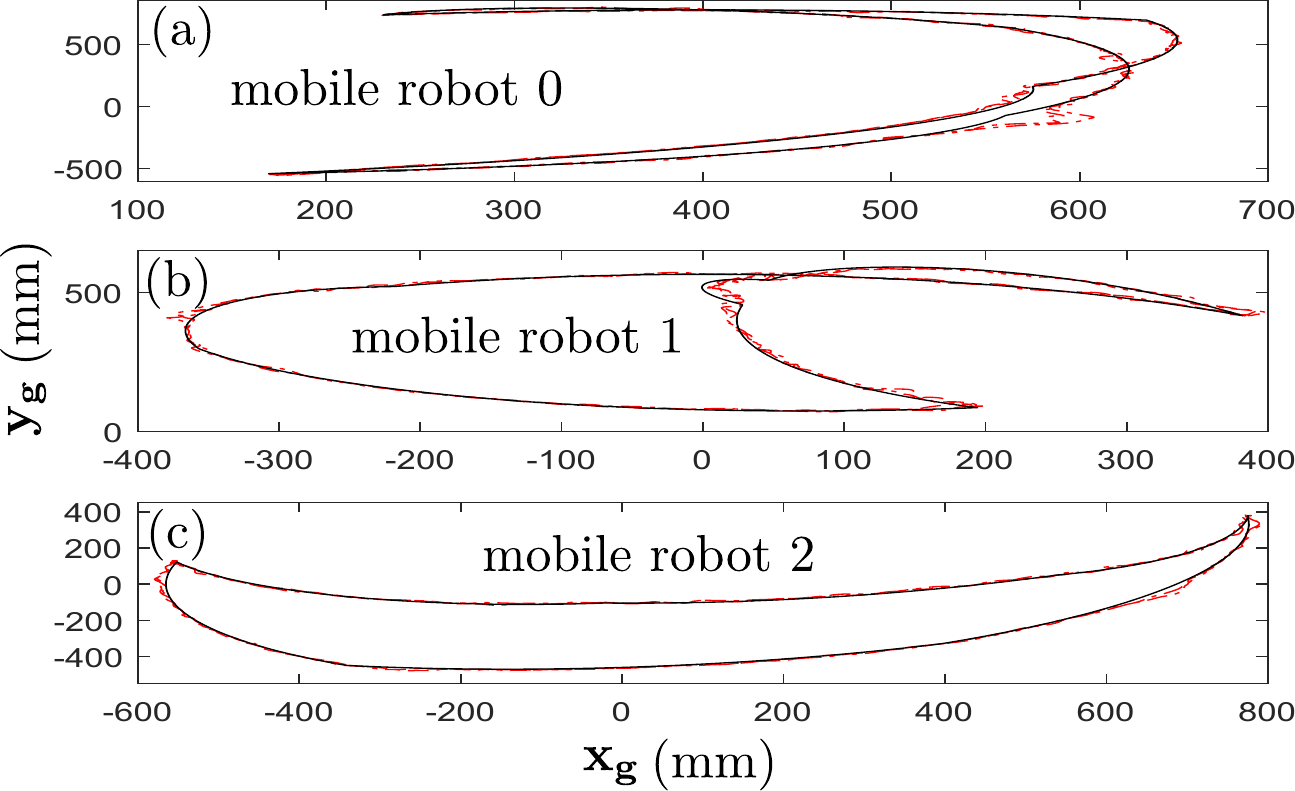}
    \caption{Motion of the three mobile bases, in the ground plane ($\Vect{x_g} - \Vect{y_g}$), generated by the optimization framework for the desired end-effector motion in Fig.~\ref{fig:data3xEETracking}.}
    \label{fig:data3mbXYTracking}
\end{figure}
The orientation of the mobile base must change to satisfy the kinematic constraints, as shown in (Fig.~\ref{fig:data3mb0TrackingTime}(a)). Fig.~\ref{fig:data3xEETracking} shows the combined simultaneous movement, along each DOF of the end-effector, with respect to time. Fig.~\ref{fig:data3mbXYTracking} shows the corresponding motion of the mobile bases with respect to the ground plane ($\Vect{x_g} - \Vect{y_g}$). RMSE were $10$, $17$, $13$~mm and $0.1$~rad for top to bottom plots, respectively, for the end-effector motion in Fig.~\ref{fig:data3xEETracking}.

Please refer to the accompanying video for results in simulation for the $4$-DOF and $6$-DOF CCMA system examples and experimental results demonstrating different kinds of motion for the two $4$-DOF CCMA prototypes.

\section{Discussion, conclusions and future Work}
\label{sec:Conclusion}
%\subsection{Discussion}
%Thus the CCMA system has the potential to be scalable to large scale systems with high mobility. This is in contrast to the parallel robots~\cite{merlet_parallel_2006}, where a similar passive kinematic chain with an end-effector is actuated with motors at the fixed base. For parallel robots, usually the number of actuators matches exactly the number of DOF of the end-effector and they have very limited workspace both in translation and orientation. 
%\subsection{Future work}
\subsection{Discussion}
Due to use of mobile bases, the robots presented in paper can be quite dexterous and have large workspaces. However, certain workspaces such as translation workspace along $\Vect{z_g}$ can be limited by the design parameters, such as link lengths, in the passive kinematic chain. In future, we aim to do the design optimization of the CCMA system for certain prescribed workspaces. Since the CCMA system effectively includes closed loop kinematic chains, treatment of singularities especially parallel singularities would be added in the optimization framework.
%to detect and avoid regions of instability. 
The multiple actuation schemes, over-actuation and increase in number of mobile bases would be exploited to provide potential solutions in the future work. %Our optimization framework is a real time simulation and control tool. %However, in order to enable closed loop feedback control over wireless networks for these untethered mobile systems, different sensing/localization systems and different communication  protocols need to be explored to maintain the real time control capability with a high bandwidth. 
As these systems are scaled, they are bound to exhibit a lot more flexibility and compliance which is not necessarily a bad outcome, as the compliance can be good for safety reasons, but its effects have to be studied and possibly included in the simulation framework for accurate prediction of the end-effector states. {Addition of on-board sensors like IMUs, cameras, lidars on top of the wheel encoder's information would enable more freedom for the CCMA system to be operated without an external motion tracking system. 
}

{The current paper dealt with a class of robots rather than evaluating a particular design of robot for a particular application. 
%In order to conclusively comment on task or application specific performance, systematic task-based design studies are needed which was not the focus of this paper. 
With an application or task specific design optimization which covers range of performance criterion such as precision, stiffness, payload and workspace, the CCMA system could in future be applicable to wide range of logistics, field robotics and service robotics tasks.} In this paper, CCMA system was limited to totally passive kinematic chains which might be limiting for the payload capacity and stiffness required for certain applications. However, apart from using mobile agents as actuators in the CCMA system, a small subset of passive joints in closed-loop kinematic chains of the CCMA system could be actuated. This increased control and actuation space has the potential to significantly increase the stiffness and payload capacity of the CCMA system.
%The play in the joints during the fabrication process can have adverse effects on the mobility of these systems as it introduces un-modelled compliances and might make the system uncontrollable. This is what we observed for the $6$~DOF prototype that due to play in the joints combined with the compliance of the plastic pipes, the fabricated system did not behave as in simulation. Therefore these issues related to proper fabrication to maintain the strict geometric arrangement of the axes and modeling of the compliance in the passive kinematic chain due to link deformation would be studied in the future work. 

\subsection{Conclusions and future work}
In this work, we have introduced a generalized concept of constrained collaborative mobile agents. These systems have the potential to be scalable and adaptable according to the different task requirements. We have presented a novel optimization framework using sensitivity analysis, which allows flexibility to test different designs, topologies of passive kinematic chain, different number of mobile agents and different actuation schemes. With results in simulation, proof of concept prototypes and experimental quantitative results, we have demonstrated the efficacy of the developed optimization tool for the simulation and kinematic control of such systems.

In this work, we presented robots actuated with omnidirectional mobile bases. In the future work, we aim to work with different mobile bases such as mobile bases with non-holonomic constraints or quadruped robots, which would lead to different instantiations of the CCMA concept.  


%%%%%%%%%%%%%%%%%%%%%%%%%%%%%%%%%%%%%%%%%%%%%%%%%%%%%%%%%%%%%%%%%%%%%%%%%%%%%%%%

%%%%%%%%%%%%%%%%%%%%%%%%%%%%%%%%%%%%%%%%%%%%%%%%%%%%%%%%%%%%%%%%%%%%%%%%%%%%%%%%

%%%%%%%%%%%%%%%%%%%%%%%%%%%%%%%%%%%%%%%%%%%%%%%%%%%%%%%%%%%%%%%%%%%%%%%%%%%%%%%%
%\section*{Appendix}

%\section{Acknowledgment}

%%%%%%%%%%%%%%%%%%%%%%%%%%%%%%%%%%%%%%%%%%%%%%%%%%%%%%%%%%%%%%%%%%%%%%%%%%%%%%%%

\bibliographystyle{IEEEtran}

\bibliography{biblio-CCR}

\end{document}